\documentclass{article}

\usepackage{arxiv}

\usepackage[utf8]{inputenc} 
\usepackage[T1]{fontenc}    
\usepackage{hyperref}       
\usepackage{url}            

\usepackage{booktabs}       

\usepackage{amsfonts}       
\usepackage{nicefrac}       
\usepackage{microtype}      
\usepackage{lipsum}
\usepackage{graphicx}
\usepackage{float}
\usepackage{amsmath}
\usepackage{cleveref}
\usepackage[backend=biber,style=ieee,sorting=none,citestyle=numeric-comp]{biblatex}
\usepackage{multicol}
\usepackage{listings}

\usepackage[ruled,linesnumbered]{algorithm2e}

\DontPrintSemicolon
\SetKwComment{Comment}{\#\ }{}
\SetKwProg{Fn}{def}{:}{}

\graphicspath{ {./img/} }

\title{Differentiable RNA Secondary Structure Extraction for Deep Learning}

\author{
 Tyler Illman \\
  School of Physics, Maths and Computing\\
  The University of Western Australia\\
  \texttt{tyler.illman@research.uwa.edu.au} \\
   \And
 Max Ward \\
  School of Physics, Maths and Computing\\
  The University of Western Australia\\
  \texttt{max.ward@uwa.edu.au} \\
  \And
 Marcell Szikszai \\
  School of Physics, Maths and Computing\\
  The University of Western Australia\\
  \texttt{marcell.szikszai@uwa.edu.au} \\
\And
 Ryan K. Krueger \\
 School of Engineering and Applied Sciences \\
 Harvard University \\
 \texttt{ryan\_krueger@g.harvard.edu} \\
}

\begin{document}

\maketitle

\begin{abstract}

Many deep learning approaches to RNA secondary structure prediction have recently been proposed. They typically output a weight matrix $W$ where $W_{ij}$ is an arbitrary weight for base $i$ pairing with base $j$. Converting this matrix to a predicted secondary structure or base-pairing probability matrix typically involves ad hoc and problematic downstream algorithms. Despite the importance of this conversion step, which we refer to as structure extraction, it has received relatively little attention in the literature. In this work, we analyze how the congruence between training and extraction methods affects prediction performance. To do this, we compare four extraction algorithms: a Nussinov-like dynamic programming method, maximum-weight graph matching and the greedy extraction algorithms used by SPOT-RNA and RiNALMo. These are evaluated on outputs from the pretrained RiNALMo model and three toy models trained in this paper: a differentiable Nussinov-like model, a binary cross-entropy (BCE) baseline, and a model that incorporates a novel symmetric doubly stochastic matrix (SDSM) normalization algorithm during training which allows it to output base-pairing probability matrices directly, without a separate extraction step. This SDSM normalization algorithm is differentiable and can be added inline to any deep learning model during training and evaluation.
We find that the performance of each extraction method depends strongly on how the corresponding model was trained. Considering the toy models themselves, the SDSM model showed the strongest overall performance: it outperformed the BCE baseline under all four extraction algorithms and produced pre-extraction outputs closest to the ground truth. These results suggest that SDSM normalization is a tractable alternative to traditional structure extraction.

\end{abstract}

\newpage

\section{Introduction}

Ribonucleic acid (RNA) plays a vital role in many biological processes. Although early research primarily focused on coding RNAs and their role in protein synthesis \cite{crick1970central}, the discovery of the broader group of non-coding RNAs (ncRNAs) has significantly expanded our understanding of the function of RNA. These ncRNAs are now known to be involved in various processes, including diseases \cite{cohen2007protein}, gene regulation \cite{serganov2007ribozymes} and catalysis \cite{doudna2002chemical}. Importantly, research has shown that the function of ncRNAs is largely governed by their secondary structure (the specific pattern of base-pairing interactions that emerge as the RNA folds) \cite{tinoco1999rna}.

Given the critical role of secondary structure in determining RNA function, accurately predicting it from the known sequence of nucleotide bases has become a central challenge in computational biology. Early methods relied on thermodynamic models such as minimum free energy (MFE) approaches pioneered by Nussinov \cite{nussinov1978algorithms, nussinov1980fast}, which uses dynamic programming (DP) to maximize the number of base pairs. This was then extended by Zuker and Stiegler \cite{zuker1981optimal}, who incorporated empirically derived nearest-neighbor energy parameters into the DP recursion to find the MFE structure. McCaskill \cite{mccaskill1990equilibrium} furthered the thermodynamic approach by computing the partition function over all possible structures, enabling the calculation of base-pairing probabilities, providing a probabilistic view of RNA folding. These methods established the thermodynamic and probabilistic foundations of RNA structure prediction upon which many subsequent approaches have been built.

More recently, researchers have attempted to use deep learning methods to directly predict structural information from RNA primary sequences. For example, models such as SPOT-RNA \cite{singh2019rna} and RiNALMo \cite{penić2024rinalmogeneralpurposernalanguage} are trained to produce a weight matrix, where each entry reflects the likelihood of a given base pair. These models 
have claimed to achieve state-of-the-art results on benchmark datasets, despite often failing to generalize and in many cases being outperformed by traditional thermodynamic algorithms \cite{szikszai2022deep, RivasProbabilisticModels2012, FlammCaveatsDeepLearning2022, SzikszaiDeepLearningRNA2026}.
Importantly, a weight matrix is an abstract, unconstrained representation of a structure that is not directly interpretable and often not compatible with downstream tasks, which require a discrete or probabilistic representation. We discuss the varying types of structure representations further in Section~\ref{sec:validity}.
 Converting a model outputted weight matrix into an interpretable structure such as a discrete or probability matrix is a separate post-processing step that we refer to as structure extraction. Unfortunately, structure extraction is often performed in a manner that is non-differentiable and entirely independent of a model's training process and objectives.

This decoupling leads to a key limitation: a model can perform well at minimizing the numerical loss between its predicted matrix and the target, yet still produce a matrix that extracts poorly into an interpretable structure representation. We believe this can be explained, at least in part, by the degree of congruence between a model's training objective and its extraction algorithm, which we call the training-extraction congruence. This congruence is the central focus of our work.

To investigate the effect of training-extraction congruence on secondary structure prediction models, we evaluate four structure extraction methods on an overfitting experiment: a Nussinov-like extraction algorithm, maximum-weight graph matching, and the greedy extraction algorithms of SPOT-RNA and RiNALMo. We first apply each method to the same set of base-pairing weight matrices generated by a pretrained version of the full RiNALMo secondary structure prediction model. We then train three proof-of-concept models: a model that uses a differentiable Nussinov-like algorithm to compute the partition function \cite{mccaskill1990equilibrium} as part of its loss, making its training objective congruent with the Nussinov-like extraction method, a binary cross-entropy (BCE) model that serves as a baseline, and finally a model that uses a novel symmetric doubly stochastic matrix (SDSM) normalization algorithm to output a probability matrix directly. 
To our knowledge, this SDSM normalized model uses the first end-to-end differentiable loss function that allows a neural network to be directly trained to output only base-pairing probability matrices.

\section{Materials and Methodology}

\subsection{Dataset and Preprocessing}

Experiments in this paper use the ArchiveII \cite{sloma2016exact} dataset, a widely adopted and experimentally validated benchmark for RNA secondary structure prediction. ArchiveII contains 3,975 sequences drawn from a diverse range of RNA families, including 5S and 16S ribosomal RNAs (rRNAs), transfer RNAs (tRNAs), signal recognition particle RNAs (SRP RNAs), ribonuclease P RNAs (RNase P), and others. These structures have been determined using comparative or experimental methods, making ArchiveII one of the most reliable publicly available datasets of RNA structures. The full breakdown of families in the ArchiveII dataset can be seen in \Cref{tab:archiveii-families} below.

\begin{table}[H]
\centering
\caption{Breakdown of RNA families in the ArchiveII dataset.}
\label{tab:archiveii-families}
\begin{tabular}{lr}
\toprule
Family & \# Sequences \\
\midrule
5S rRNA         & 1,283 \\
SRP RNA         &   928 \\
tRNA            &   557 \\
tmRNA           &   462 \\
RNase P RNA     &   454 \\
16S rRNA        &   110 \\
Group I Intron  &    98 \\
Telomerase RNA  &    37 \\
23S rRNA        &    35 \\
Group II Intron &    11 \\
\midrule
\textbf{Total}  & \textbf{3,975} \\
\bottomrule
\end{tabular}
\end{table}

Due to numerical stability constraints of the differential folding used in the Nussinov-like model, all 3 models in this paper were trained only on sequences shorter than 200 nucleotides and any remaining RNA families with fewer than 15 sequences were excluded from the training/testing set. Fortunately, the ArchiveII dataset is heavily skewed toward shorter RNAs, largely due to the prevalence of compact families such as tRNA, SRP RNA, and 5S rRNA. Note that in some cases, such as 16S and 23S rRNA, long sequences are divided into independent folding domains to simplify the prediction task \cite{mathews1999expanded}.

The dataset was randomly split into training and test sets using an 80/20 ratio respectively. This split results in homologous sequences appearing in both sets, the deficiencies of which have been extensively documented in existing literature~\cite{szikszai2022deep, RivasProbabilisticModels2012, FlammCaveatsDeepLearning2022, SzikszaiDeepLearningRNA2026}. Normally, when evaluating the real-world performance of single-sequence models like those presented here, the most important and fair benchmarks examine the performance on novel or synthetic RNAs~\cite{SzikszaiDeepLearningRNA2026}. For molecules with known homology to existing RNA families, or sequences where other homologous sequences can be aligned, single-sequence methods should not be used~\cite{szikszai2022deep, RivasProbabilisticModels2012, FlammCaveatsDeepLearning2022, MagnusRNAHubAutomated2025, SzikszaiDeepLearningRNA2026}. However, it is important to note that the goal of this research is not to build a large state-of-the-art model that generalizes across RNA families rather, our aim is to prove the importance of training-extraction congruence and argue the potential value of SDSM normalization. As a result, an over-fitting experiment where the same RNA families appear in both the training and testing set can sufficiently address our questions. Having said that, we do not address how improved congruence on over-fitting experiments or SDSM normalization impacts generalization. Since there is no evidence that any existing deep learning model can sufficiently generalize without being heavily constrained via thermodynamic integration~\cite{SatoRNA2021, SzikszaiDeepLearningRNA2026}, we believe that comprehensively addressing this point is not currently tractable until significant advancements are made in the field.

Thus, this training/test split can be considered an over-fitting experiment as RNA families appear in both the training and testing set. A breakdown of the train/test split can be seen in \Cref{tab:dataset-type-breakdown} below.

\begin{table}[H]
\centering
\caption{Breakdown of RNA structure types in the train and test datasets.}
\label{tab:dataset-type-breakdown}
\begin{tabular}{lrr}
\toprule
RNA Type & Train Set & Test Set \\
\midrule
5S rRNA  & 1,018 & 265 \\
SRP RNA  &   411 & 114 \\
tRNA     &   461 &  96 \\
\midrule
\textbf{Total} & \textbf{1,890} & \textbf{475} \\
\bottomrule
\end{tabular}
\end{table}

RNA sequences were one-hot encoded into binary matrices of shape $L \times 4$, where $L$ is the sequence length. Corresponding secondary structures were converted into adjacency matrices of shape $L \times L$, where position $(i, j)$ was set to 1 if bases $i$ and $j$ form a base pair. For unpaired bases, the diagonal entry $(i, i)$ was set to 1. Only the upper triangle of each matrix was used for training, consistent with prior work \cite{singh2019rna}.

\subsection{Interpretable Secondary Structure Representations}
\label{sec:validity}

Throughout this work we distinguish between three different secondary structure representations: a base-pairing weight matrix, a probability matrix, and a discrete structure. Most deep learning models are trained to output a weight matrix, despite the fact that most downstream tasks and tools, such as evaluation, tertiary structure prediction tools \cite{PopendaAutomated3D2012}, and analysis tools \cite{LorenzViennaRNAPackage2011} require a directly interpretable representation such as a probability matrix or a discrete structure.

A base-pairing weight matrix is an $L \times L$ matrix $W=(w_{ij})$, where each entry $w_{ij}$ assigns a score to bases $i$ and $j$ pairing. These scores indicate the relative preference of the model for different pairings, but do not necessarily have a probabilistic interpretation. Unless explicitly constrained, the matrix may be asymmetric, its rows and columns need not sum to one, and a single base may simultaneously receive high weights for several conflicting pairing partners. Even when the entries are bounded to $[0,1]$ using an activation such as a sigmoid, they are typically predicted independently and therefore do not form a probability distribution over the possible states of each base. Consequently, a weight matrix cannot generally be interpreted directly as a secondary structure.

A discrete secondary structure instead assigns each base exactly one state: it is either paired with one other base or remains unpaired. We represent this as a binary $L \times L$ adjacency matrix $A=(a_{ij})$, where $a_{ij}=a_{ji}=1$ when bases $i$ and $j$ are paired and $a_{ii}=1$ when base $i$ is unpaired. All other entries are zero. The resulting matrix is symmetric, with every row and column summing to one.

A base-pairing probability matrix provides the probabilistic analogue of this representation. We denote this matrix by $P=(p_{ij})$. Each off-diagonal entry $p_{ij}$ gives the probability that bases $i$ and $j$ are paired, while the diagonal entry $p_{ii}$ gives the probability that base $i$ remains unpaired. The possible states of each base therefore form a probability distribution, requiring the matrix to be non-negative and for every row and column to sum to one. Pairing probabilities are also symmetric, since the probability of base $i$ pairing with base $j$ is the same as the probability of base $j$ pairing with base $i$.

We refer to probability matrices and discrete structures as \emph{interpretable representations} because they directly describe the pairing state of each nucleotide and can therefore be interpreted and used by downstream methods directly. Both representations are non-negative and symmetric, with every row and column summing to one and the unpaired state represented on the diagonal. Matrices satisfying these properties are also precisely symmetric doubly stochastic matrices (SDSMs), which is discussed further in Section~\ref{sec:sdsm-model}. An unconstrained weight matrix generally does not satisfy these properties and must therefore be converted into an interpretable representation through an additional procedure, which we refer to throughout this work as \emph{structure extraction}. The SDSM normalization introduced later instead enforces these properties differentiably, allowing a model to be trained end-to-end to produce a base-pairing probability matrix without requiring a separate extraction step.

\subsection{Secondary Structure Extraction Methods}
\label{sec:extraction-methods}

As mentioned previously, a structure extraction method is an algorithm that converts a model's base-pairing weight matrix into an interpretable secondary structure representation, as defined in Section~\ref{sec:validity}. We evaluate four such methods, each takes an $L \times L$ weight matrix, where entry $(i, j)$ scores the pairing of bases $i$ and $j$, and returns a discrete secondary structure. 

\subsubsection{RiNALMo's Structure Extraction Algorithm}

RiNALMo \cite{penić2024rinalmogeneralpurposernalanguage} uses a greedy algorithm to extract a discrete secondary structure from base-pairing weight matrices. Before applying this selection procedure, several pre-processing steps are used to pre-process the matrix. First, non-canonical base pairs are masked out by zeroing entries that do not correspond to canonical or wobble base pairings. Next, a sharp loop filter eliminates pairs where $|i - j| < 4$. Finally, all remaining entries below a fixed confidence threshold (typically 0.5) are set to zero.

The pre-processed matrix is then processed greedily. At each iteration, the highest-weight base pair $(i, j)$ is selected and added to the structure. All other pairings involving base $i$ or base $j$ are eliminated by zeroing their corresponding rows and columns. This continues until no valid entries remain. While this method is fast and guarantees that no base is paired more than once, it is not guaranteed to find the globally optimal pairing configuration.

\subsubsection{SPOT-RNA's Structure Extraction Algorithm}

SPOT-RNA \cite{singh2019rna} also extracts a discrete secondary structure using a greedy extraction algorithm. Similar to RiNALMo, invalid positions and pairs are first masked out based on one-hot encoding. By default, SPOT-RNA allows non-canonical base pairs, and this default behavior was retained in our experiments. A fixed threshold (0.335 by default) is then applied to remove low-confidence entries.

Unlike RiNALMo, SPOT-RNA initially allows a single base to be paired with more than one other base, known as a multiplet. All valid base pairs above the threshold are initially selected, even if this results in some bases being paired more than once. Conflicts are later resolved by keeping only the highest-weighted pair for each base. The final structure consists of the remaining mutually exclusive pairs. This delayed pruning strategy allows for more globally optimal decisions during extraction.

\subsubsection{Maximum Weight Matching Extraction Algorithm}

Maximum weight matching is a problem in graph theory that tries to find a graph matching with maximum sum of weights. For our purposes, we reinterpret the weight matrix as a weighted graph, where each nucleotide is represented as a node and the confidence score between two bases defines the weight of the edge between them. An additional ``unpaired'' node is used to capture the weight of bases being unpaired (i.e. the matrix diagonal). Edges are only added between bases if they are canonical pairs (including wobble pairs) and satisfy a minimum loop length (set to 3 by default). The algorithm then solves a maximum weight matching problem using Edmonds' Blossom algorithm \cite{galil1986efficient} as implemented in the NetworkX package \cite{SciPyProceedings_11}. This ensures that no nucleotide is paired more than once and that the set of base pairs has maximum total weight under the given constraints.

\subsubsection{Nussinov-like Extraction Algorithm}
\label{sec:nussinov_extraction}

This extraction method adapts Nussinov's classic dynamic programming algorithm \cite{nussinov1980fast} to operate on model-predicted pairing scores instead of thermodynamic energies. Given an arbitrary weight matrix $A=(a_{ij})$, where each entry $a_{ij}$ reflects the score assigned to bases $i$ and $j$ pairing and each diagonal entry $a_{ii}$ reflects the score assigned to base $i$ remaining unpaired, the algorithm constructs a dynamic programming matrix $S$. Each entry $S[i,j]$ stores the maximum achievable score for a nested secondary structure spanning the subsequence from base $i$ to base $j$.

The recurrence relation used to populate $S$ is given in \Cref{eq:nussinov_rec}, with $S[i,j]=0$ when $j<i$. It considers two cases for the leftmost base $i$: leaving $i$ unpaired, or pairing $i$ with a valid partner $k$. If $i$ remains unpaired, the optimal score is given by the score of the remaining interval $S[i+1,j]$ together with the unpaired weight $a_{ii}$. If $i$ is paired with $k$, the interval is divided into the two independent sub-intervals $[i+1,k-1]$ and $[k+1,j]$, with the pairing contributing weight $a_{ik}$. The optimal score is obtained by taking the maximum over the unpaired case and all valid pairing partners $k$:

\begin{equation}
\label{eq:nussinov_rec}
S[i,j]
=
\max \left\{
S[i+1,j] + a_{ii},
\;
\max_{\substack{
k=i+\mathrm{min\_loop}+1,\ldots,j \\
(i,k)\text{ is a valid pair}
}}
\left(
S[i+1,k-1]
+
S[k+1,j]
+
a_{ik}
\right)
\right\}.
\end{equation}

After the matrix is filled, a traceback is performed to recover the optimal structure. The traceback ensures that only non-crossing, symmetric base pairs are included, thereby enforcing structural constraints such as canonical pairing rules, a minimum loop length, and the exclusion of pseudoknots. This guarantees the extracted structure is both discrete and nested. It is also guaranteed to be the maximum-weight set of base pairs under these constraints.

\subsection{Extraction Evaluation Using Pretrained RiNALMo Outputs}

To assess the performance of the different structure extraction methods independently of model training, we first evaluated each algorithm on the same set of base-pairing weight matrices generated by the full pretrained RiNALMo model \cite{penić2024rinalmogeneralpurposernalanguage}. This set of base-pairing weight matrices was generated using the full ArchiveII data set.

All four extraction methods: RiNALMo's, SPOT-RNA's, Maximum Weight Matching, and the Nussinov-like extraction were applied to these matrices to generate discrete secondary structures. The predicted structures were then compared against the ground truth annotations provided in ArchiveII using F1 score as the evaluation metric. Again, it is important to note that analysis of the pretrained RiNALMo model's performance \textbf{cannot} be compared directly against the performance of the models trained in this paper. RiNALMo is a far more sophisticated model, trained on a larger, more diverse training set, as opposed to the small, proof-of-concept, overfitting models trained in this paper.

\subsection{Model Architectures and Implementation}

In addition to using RiNALMo, we trained our own models to demonstrate the effect of training-extraction congruence. All models trained in this paper share an identical fully connected multi-layer perceptron (MLP) architecture to produce a $200 \times 200$ base-pairing matrix from a one-hot encoded RNA sequence. All sequences are padded to a fixed length of 200 nucleotides to ensure compatibility with the model input. A max sequence length of 200 bases was selected to improve numerical stability during training, this was particularly important for the Nussinov-like model \cite{krueger2025jax}. The only differences between models were in their final layer activation functions and loss functions applied during training. A breakdown of model architecture can be found in Appendix~\ref{app:model-architectures}.

The general model architecture consisted of a flattened input layer representing a $200 \times 4$ one-hot encoded sequence, followed by three fully connected hidden layers of size 1024, 512 and 256 respectively, each using ReLU activations. The final output layer produces a $200 \times 200$ matrix of raw scores, reshaped from a vector of size 40,000. All models were implemented in JAX \cite{jax2018github} and trained using the Flax library \cite{flax2020github}. Training was performed on Google Colab Pro with an NVIDIA L4 GPU. We used the Adam optimizer \cite{kingma2017adammethodstochasticoptimization} with a learning rate of $1 \times 10^{-3}$, a batch size of 16, and 32 training epochs. Code and evaluation scripts are available at \href{https://github.com/TylerIllman/RNA-Secondary-Structure-Extraction}{github.com/TylerIllman/RNA-Secondary-Structure-Extraction}.

\subsubsection{Nussinov-like Model}

The Nussinov-like model was designed to be highly congruent with the Nussinov-like extraction function described in Section~\ref{sec:nussinov_extraction}. It predicts a $200 \times 200$ matrix $E = (e_{ij})$ of unitless base-pairing pseudo-free energies, where $e_{ij}$ is the predicted pseudo-energy of bases $i$ and $j$ pairing, $e_{ii}$ represents the predicted pseudo-energy of base $i$ remaining unpaired. Importantly, these \emph{pseudo-free energies} are not real free energies, they are predicted, unitless scores that are treated as energies and converted to Boltzmann weights with the gas constant and temperature terms omitted from the formulation. A \texttt{tanh} activation bounds these energies to $[-1,1]$. This range suits the free-energy interpretation by allowing favorable interactions to receive negative energies while also preventing extreme Boltzmann weights. Since the partition function accumulates products and sums over a combinatorially large number of structures, numerical stability was further improved using a per-nucleotide scaling factor of $0.5$, a maximum sequence length of 200 nucleotides, and gradient clipping during optimization. 

The predicted pseudo-energies are converted into a Boltzmann weight matrix $B=(b_{ij})$, where each entry $b_{ij}$ is defined as:

\begin{equation}
b_{ij} = \exp(-e_{ij}),
\label{eq:boltzmann_weight}
\end{equation}

and the gas constant and temperature ($k_BT$) are omitted. Thus, for a valid secondary structure $s$, its Boltzmann weight $w(s)$ is defined as:

\begin{equation}
w(s) = \prod_{(i,j)\in s} b_{ij},
\label{eq:structure_weight}
\end{equation}

where $b_{ij}$ is the Boltzmann weight of base $i$ and base $j$ pairing. Note, the Nussinov-like model could equally have been trained to directly output a Boltzmann weight matrix, removing the need for this conversion step. However, we opted to include it to match the thermodynamic convention \cite{mccaskill1990equilibrium}.

Let $\mathcal{S}$ denote the set of all valid secondary structures for the sequence, and let $s \in \mathcal{S}$ denote a particular secondary structure. The global partition function $Z$ and the probability of structure $s$ ($p(s)$), are then defined as:

\begin{equation}
Z = \sum_{s\in\mathcal{S}} w(s),
\label{eq:partition_function}
\end{equation}

\begin{equation}
p(s) = \frac{w(s)}{Z}.
\label{eq:structure_probability}
\end{equation}

We are able to compute $Z$ using a modified version of the classic Nussinov recurrence. Unlike the original discrete approach, which selects the optimal structure via a \texttt{max} operation, this formulation sums over all valid structures. The recurrence used to compute the partition function can be seen below in \Cref{eq:diff_nussinov_rec}.

\begin{equation}
Z[i,j] =
Z[i+1,j] \times b_{ii}
+ \sum_{k=i+\mathrm{min\_loop}+1}^{j}
\left(
Z[i+1,k-1] \times Z[k+1,j] \times b_{ik}
\right).
\label{eq:diff_nussinov_rec}
\end{equation}

In this formulation, the total partition function $Z[i,j]$ sums over all possible RNA secondary structures within the span from base $i$ to $j$. The first term corresponds to the case where base $i$ is unpaired, contributing the Boltzmann weight $b_{ii}$ which is drawn from the diagonal of the predicted weight matrix and the partition function of the remaining span $[i+1,j]$.

The second term considers all valid pairing partners $k$ that satisfy the minimum loop length constraint. For each valid base pair $(i,k)$, the contribution is the product of the partition functions over the two resulting subintervals: $[i+1,k-1]$ and $[k+1,j]$ multiplied by the Boltzmann weight $b_{ik}$. Because the recurrence is computed in exponentiated space, this product effectively sums the energies of all components in log space. By summing over all such possibilities, this recurrence accumulates the total Boltzmann-weighted contribution of all valid substructures within the interval $[i,j]$, enabling end-to-end differentiability.

The Nussinov-like model was trained to maximize the probability $p(s_{\mathrm{true}})$ of the ground-truth structure $s_{\mathrm{true}}$ given the sequence. This approach is closely related to recent differentiable folding frameworks that implement partition function--based formulations for RNA secondary structure optimization and prediction tasks \cite{matthies2024differentiable, krueger2025jax}. Specifically, we defined the loss as the negative log-likelihood of the ground-truth structure under the predicted Boltzmann distribution, where $w(s_{\mathrm{true}})$ is the Boltzmann weight of the ground-truth structure and $Z$ is the global partition function, with additional regularization:

\begin{equation}
\mathcal{L}
=
-\left(
\log w(s_{\mathrm{true}}) - \log Z
\right)
+
\lambda \| \theta \|_2^2.
\label{eq:nussinov_loss}
\end{equation}

Maximizing $\log w(s_{\mathrm{true}})-\log Z$ increases the probability assigned to the ground-truth structure relative to all other valid structures. The final term applies $L_2$ regularization to the model parameters $\theta$ for numerical stability, controlled by the hyperparameter $\lambda$. However, this loss function does not constrain the model output to be an interpretable secondary structure representation. Instead, the model outputs a unitless pseudo-free energy matrix $E=(e_{ij})$, which is converted into a Boltzmann weight matrix $B=(b_{ij})$ and requires a downstream structure extraction step to obtain an interpretable structure.

\subsubsection{Binary Cross-Entropy Baseline Model}
\label{sec:cbbce-model}

As a baseline, we trained a model that treats secondary structure prediction as a per-entry binary classification problem, predicting for each entry of the matrix whether it is active. The model produces a $200 \times 200$ matrix of raw scores, which is symmetrized and passed through a sigmoid activation to give a predicted weight matrix $A=(a_{ij})$ with entries $a_{ij} \in [0,1]$. Binary cross-entropy is the standard loss for this kind of classification, and it is the objective used by nearly all existing deep learning predictors: both SPOT-RNA \cite{singh2019rna} and RiNALMo \cite{penić2024rinalmogeneralpurposernalanguage} predict per-entry probabilities through a sigmoid and train against the known structure with a cross-entropy objective. This makes a binary cross-entropy model a natural and representative baseline. Because the sigmoid treats each entry independently, the predicted values for a given base are not constrained to sum to one, so the output is a base-pairing weight matrix and is not an interpretable representation in the sense of Section~\ref{sec:validity}. Thus it relies on a downstream extraction step. 

Importantly, each entry in the predicted matrix falls into one of two classes: a predicted interaction or a non-predicted interaction. These two classes are severely imbalanced. Because each base pairs with at most one partner, the matrix is sparse and dominated by non-predicted interactions, which far outnumber the predicted interactions. To stop this majority class from overwhelming training, we weight the loss using the class-balanced scheme of Cui et al. \cite{CuiClassBalancedLoss2019}. This method weights each class by its effective number of samples, $C_m$, a saturating count that grows more slowly than the raw count $m$. Weighting by $C_m$ rather than the raw count down-weights the dominant non-predicted interaction class relative to the rare predicted entries class, as seen below:

\begin{equation}
    C_{m} = \frac{1 - \beta^{m}}{1 - \beta},
    \qquad
    \beta = \frac{N - 1}{N},
    \qquad
    N = n_{+} + n_{-},
\end{equation}

\begin{equation}
    \alpha_{+} = \frac{1}{C_{n_{+}}} = \frac{1 - \beta}{1 - \beta^{\,n_{+}}},
    \qquad
    \alpha_{-} = \frac{1}{C_{n_{-}}} = \frac{1 - \beta}{1 - \beta^{\,n_{-}}}.
\end{equation}

The loss applies one class weight to each class and sums the per-entry log-likelihoods. Denoting the positive, predicted interactions class ($\mathcal{P}$) as the entries with target value one and the negative, non-predicted interactions class ($\mathcal{N}$) as those with target value zero, the loss then has the compact form:

\begin{equation}
    \mathcal{L} =
    -\left(
        \alpha_{+} \sum_{(i,j) \in \mathcal{P}} \log a_{ij}
        + \alpha_{-} \sum_{(i,j) \in \mathcal{N}} \log\!\left(1 - a_{ij}\right)
    \right)
    + \lambda \lVert \theta \rVert_2^2 .
\end{equation}

Equivalently, using the ground-truth target matrix $T=(t_{ij})$ and the validity mask $M=(m_{ij})$ to select these entries, the loss is written as:

\begin{equation}
    \mathcal{L} =
    -\left(
        \alpha_{+} \sum_{i,j} m_{ij}\, t_{ij} \log a_{ij}
        + \alpha_{-} \sum_{i,j} m_{ij}\,(1 - t_{ij}) \log\!\left(1 - a_{ij}\right)
    \right)
    + \lambda \lVert \theta \rVert_2^2 .
\end{equation}

Here $T=(t_{ij})$ is the ground-truth target matrix, in which an entry $t_{ij}$ is set to one to mark a base pair, $t_{ii}$ is set to one to mark an unpaired base, and every other entry is set to zero. The positive class is therefore all entries equal to one. $A=(a_{ij})$ is the predicted weight matrix after the sigmoid, with entries clipped to $[\epsilon,1-\epsilon]$ where $\epsilon=10^{-6}$ for numerical stability, and $M=(m_{ij})$ restricts the loss to valid, unpadded entries in the upper triangle. For each training matrix, $N$ is the number of valid entries, and $n_{+}$ and $n_{-}$ are the counts of entries equal to one and zero within it.

Although the loss sums over entries rather than averaging, the class-balanced weights implicitly normalize it. For the rare positive class, $\alpha_+ \approx 1/n_+$, so the weighted sum is approximately the mean log-likelihood over positive entries, and similarly for the negative class. The loss is therefore roughly independent of sequence length and balanced between the two classes. The final term applies $L_2$ regularization to the parameters $\theta$, controlled by $\lambda$. Because the BCE objective treats each matrix entry independently, it does not impose the structural constraints required for an interpretable secondary structure representation. In particular, the predicted weight matrix is not required to be symmetric or doubly stochastic, and therefore requires a downstream structure extraction step to obtain an interpretable structure.

\subsubsection{Symmetric Doubly Stochastic Matrix (SDSM) Model}
\label{sec:sdsm-model}

Unlike the two preceding models, the SDSM model is designed to directly output an interpretable representation, removing the need for a downstream extraction step. A matrix $P = (p_{ij})$ is symmetric when $p_{ij}=p_{ji}$ for every pair of indices, meaning that the score assigned to bases $i$ and $j$ is independent of their ordering. A matrix is stochastic when its entries are non-negative and each row sums to one. If both its rows and columns sum to one, it is doubly stochastic. A matrix satisfying both symmetry and double stochasticity is therefore called a symmetric doubly stochastic matrix (SDSM).

An SDSM naturally represents a base-pairing probability matrix. Each off-diagonal entry $p_{ij}$ gives the probability that bases $i$ and $j$ pair, while each diagonal entry $p_{ii}$ gives the probability that base $i$ remains unpaired. Symmetry ensures that both bases assign the same probability to their shared pairing, while double stochasticity ensures that the possible pairing states associated with each base sum to one, allowing them to be interpreted as a probability distribution. By constraining its output to be an SDSM, the model produces a base-pairing probability matrix by design.

To produce an SDSM we apply a normalization procedure, which we call SDSM normalization, to the raw model output. The output is first symmetrized and then passed through a softplus function, $\mathrm{softplus}(x) = \log(1 + e^{x})$, which maps the scores to strictly positive values, as required for the normalization that follows. The positive matrix is then normalized to be doubly stochastic using a modified Sinkhorn-Knopp algorithm \cite{sinkhorn1964relationship}. The standard Sinkhorn-Knopp algorithm normalizes a strictly positive matrix to be doubly stochastic by repeatedly rescaling its rows and columns until each sums to one, and is guaranteed to converge for any strictly positive matrix \cite{sinkhorn1964relationship}.

As this does not preserve symmetry, we add a symmetrization step to each iteration. Every iteration divides each column by its sum, then divides each row by its sum, and finally averages the matrix with its transpose.

This ensures that the matrix is driven toward double stochasticity while remaining symmetric throughout. To keep the procedure differentiable under JAX, it runs for a fixed number of iterations rather than looping until convergence. However, convergence is still tracked: once all row and column sums lie within a tolerance of $10^{-8}$ of one, further updates are masked and the matrix is left unchanged for the remaining iterations so that modifications are not made on an already converged matrix. The full implementation is given in Appendix~\ref{app:sdsm-code}.

The resulting matrix is a base-pairing probability matrix by design. It is compared to the ground-truth structure using the same class-balanced binary cross-entropy loss as the baseline model, with $\mathcal{P}$ and $\mathcal{N}$ again denoting the entries whose target value is one and zero respectively:

\begin{equation}
\mathcal{L} =
-\left(
    \alpha_{+} \sum_{(i,j) \in \mathcal{P}} \log p_{ij}
    + \alpha_{-} \sum_{(i,j) \in \mathcal{N}} \log\!\left(1 - p_{ij}\right)
\right)
+ \lambda \lVert \theta \rVert_2^2 .
\end{equation}

The class weights $\alpha_{+}$ and $\alpha_{-}$ are the effective-number weights from Section~\ref{sec:cbbce-model}. The only difference from the baseline is that $P$ is now the doubly stochastic output of SDSM normalization rather than an independently predicted sigmoid output applied to each entry. Because this output is already a probability matrix, no extraction is required to interpret it, though for consistency with the other models we still apply the four extraction algorithms to its output when reporting results.

\subsection{Treatment of Pseudoknots}
\label{sec:pseudoknots}

Although most RNA secondary structure prediction methods assume that structures are fully nested, some biologically important RNAs contain pseudoknots. A pseudoknot is a non-nested base pair. More formally, given two base pairs $(i,j)$ and $(k,l)$, a pseudoknot occurs when their indices interleave such that $i < k < j < l$. In contrast, a nested secondary structure contains only non-crossing base pairs. Thus, allowing pseudoknots prevents the use of the dynamic programming recurrences used for nested structures, requiring more general algorithms that can become computationally expensive.

Pseudoknots contribute to the folding, stability, and function of specific RNA molecules, making them an important consideration when developing comprehensive structure prediction methods \cite{RivasDynamicProgramming1999}. However, 
Pseudoknots only account for 1.4\% of all RNA base-pairing interactions \cite{mathews1999expanded}, and many of the widely used structure prediction methods focus only on nested structures \cite{SaccoMachinelearning2026}.
Moreover, including pseudoknots complicates RNA folding as the recursive decomposition used by classical dynamic programming algorithms are no longer sufficient, making the computational complexity significantly greater \cite{LyngsoRNAPseudoknot2000,LyngsoComplexityPseudoknot2004}.

The models and extraction methods considered in this work differ in their ability to represent pseudoknotted structures. The nested recursion used by the Nussinov-inspired training objective permits only nested base pairs and therefore excludes pseudoknots. In contrast, the BCE and SDSM training objectives do not impose a nesting constraint, allowing their output matrices to represent pseudoknots. The pairwise outputs of the pretrained RiNALMo model are similarly not restricted to nested structures. 

Pseudoknot support also differs among the extraction methods. Nussinov extraction again produces only nested structures, whereas Graph Matching, RiNALMo extraction, and SPOT-RNA extraction support pseudoknotted structures. Consequently, applying Nussinov extraction excludes pseudoknots regardless of the originating model, while the other extraction methods can retain pseudoknotted interactions represented in the model output. Moreover, we do not assess each model and extraction method's ability to predict or extract pseudoknots in this work.

\subsection{Evaluation Metrics}

We used two standard metrics to assess model performance at different stages of the pipeline: mean squared error (MSE) to evaluate output matrices prior to extraction and F1 score to measure the accuracy of the final extracted secondary structures. Specifically, we use the relaxed F1 score recommended by Mathews \cite{MathewsBenchmarkRNA2019}, in which a predicted base pair is counted as correct if it matches the ground-truth pair exactly or after a one-position shift in one of the four cardinal directions (up, down, left, or right).

Statistical significance testing is important to determine whether the difference in prediction accuracy between two methods is greater than would be expected by chance. A paired $t$-test is a common approach for this purpose, as it assesses whether the mean paired difference in performance across sequences is significantly different from zero \cite{MathewsBenchmarkRNA2019}. In this work, however, we are primarily interested in whether one method consistently achieves 
a higher pairwise F1 score,
rather than the magnitude of the mean F1-score difference. We therefore use a custom bootstrapping permutation test based on the number of sequence-level wins.
This test works by recording whether one extraction method achieved a higher F1 score than another for a particular secondary structure, known as a win. The observed number of wins was compared to a null distribution generated by randomly swapping the extraction method assignments for each sequence across $N = 10{,}000$ permutations. In each permutation, we counted how often Method A outperformed Method B under random assignment.

Formally, the $p$-value was computed as:
\begin{equation}
p = \frac{1}{N} \sum_{j=1}^N \mathbf{1}\{w_j \geq w_{\text{obs}}\},
\label{eq:perm_test_pval}
\end{equation}
where $w_j$ is the number of wins in permutation $j$ and $w_{\text{obs}}$ is the actual observed number of wins. If $p < 0.05$, we rejected the null hypothesis and concluded that the difference in performance was statistically significant.

\section{Results}

\subsection{Extraction Method Performance on RiNALMo Outputs}

\begin{table}[H]
\centering
\caption{Mean F1 score by RNA family for each structure extraction method (from pretrained RiNALMo model outputs).}
\label{tab:f1-scores-by-family-rinalmo}
\begin{tabular}{lcccc}
\toprule
Family & RiNALMo F1 & Graph Matching F1 & Nussinov F1 & SPOT-RNA F1 \\
\midrule
16S rRNA        & 0.5690 & 0.5855 & 0.5990 & 0.5787 \\
23S rRNA        & 0.6337 & 0.6587 & 0.6636 & 0.6515 \\
5S rRNA         & 0.9275 & 0.8942 & 0.9308 & 0.9224 \\
RNase P RNA     & 0.8820 & 0.8851 & 0.8686 & 0.8874 \\
Group I Intron  & 0.8028 & 0.8055 & 0.7844 & 0.8083 \\
SRP RNA         & 0.7973 & 0.8015 & 0.8100 & 0.7921 \\
tRNA            & 0.9921 & 0.9798 & 0.9935 & 0.9649 \\
Telomerase RNA  & 0.3175 & 0.3479 & 0.2976 & 0.3405 \\
tmRNA           & 0.8088 & 0.8267 & 0.7134 & 0.8256 \\
\midrule
\textbf{Mean}   & \textbf{0.8713} & \textbf{0.8626} & \textbf{0.8630} & \textbf{0.8675} \\
\bottomrule
\end{tabular}
\end{table}

When applied to the same set of weight matrices generated by the existing pretrained RiNALMo model on the full ArchiveII set, all four extraction methods achieved high F1 scores with only marginal differences between them. As shown in \Cref{tab:f1-scores-by-family-rinalmo}, RiNALMo extraction achieved the highest overall mean F1 score (0.8713), followed closely by SPOT-RNA (0.8675), Nussinov (0.8630), and Graph Matching (0.8626). This indicates that the pretrained RiNALMo outputs were relatively insensitive to the choice of extraction method. The bootstrapping permutation tests confirmed that RiNALMo extraction significantly outperformed Graph Matching and SPOT-RNA, while there was insufficient evidence to conclude a difference between RiNALMo and Nussinov extraction. Nussinov also significantly outperformed both Graph Matching and SPOT-RNA. See Appendix~\ref{app:rinalmo-permutation-histograms} for full statistical significance testing.

It is important to note that these results are not directly comparable to those of the proof-of-concept models trained in this paper. RiNALMo is a much larger model trained on diverse datasets with a substantially more sophisticated architecture and optimization procedure. Thus, these results serve primarily as contextual background for the training-extraction congruence analysis that follows.

\subsection{Nussinov-like Model Results}

\begin{table}[H]
\centering
\caption{Mean F1 score by RNA family for each structure extraction method using the Nussinov-like model.}
\label{tab:f1-by-family-nussinov}
\begin{tabular}{lcccc}
\toprule
Family & RiNALMo F1 & Graph Matching F1 & Nussinov F1 & SPOT-RNA F1 \\
\midrule
5S rRNA  & 0.6944 & 0.9648 & 0.8787 & 0.7151 \\
SRP RNA  & 0.5587 & 0.7747 & 0.7561 & 0.5551 \\
tRNA     & 0.6681 & 0.9500 & 0.8586 & 0.7189 \\
\midrule
\textbf{Mean} & \textbf{0.6524} & \textbf{0.9163} & \textbf{0.8451} & \textbf{0.6774} \\
\bottomrule
\end{tabular}
\end{table}

The differentiable Nussinov-like model produced clear differences in performance across extraction methods. As shown in \Cref{tab:f1-by-family-nussinov}, Graph Matching achieved the highest overall mean F1 score (0.9163), followed by Nussinov extraction (0.8451). In contrast, the two greedy extraction methods performed substantially worse, with SPOT-RNA achieving a mean F1 of 0.6774 and RiNALMo achieving 0.6524. These differences were supported by the bootstrapping permutation tests, which showed that Graph Matching significantly outperformed all other extraction methods, Nussinov extraction significantly outperformed both greedy methods and SPOT-RNA also significantly outperformed RiNALMo. The full statistical comparisons are provided in Appendix~\ref{app:nussinov-permutation-histograms}.

\subsection{Binary Cross-Entropy Baseline Model Results}

\begin{table}[H]
\centering
\caption{Mean F1 score by RNA family for each structure extraction method using the class-balanced binary cross-entropy model.}
\label{tab:f1-by-family-cbbce}
\begin{tabular}{lcccc}
\toprule
Family & RiNALMo F1 & Graph Matching F1 & Nussinov F1 & SPOT-RNA F1 \\
\midrule
5S rRNA  & 0.7615 & 0.6765 & 0.8809 & 0.9302 \\
SRP RNA  & 0.6152 & 0.2755 & 0.7571 & 0.7099 \\
tRNA     & 0.7038 & 0.7783 & 0.8641 & 0.8972 \\
\midrule
\textbf{Mean} & \textbf{0.7147} & \textbf{0.6008} & \textbf{0.8478} & \textbf{0.8707} \\
\bottomrule
\end{tabular}
\end{table}

The class-balanced binary cross-entropy model produced a different ordering of extraction methods to the Nussinov-like model. As shown in \Cref{tab:f1-by-family-cbbce}, SPOT-RNA achieved the highest overall mean F1 score (0.8707), followed by Nussinov extraction (0.8478) and RiNALMo (0.7147), while Graph Matching performed worst (0.6008). The bootstrapping permutation tests showed that SPOT-RNA significantly outperformed all other extraction methods, Nussinov extraction significantly outperformed both Graph Matching and RiNALMo, and RiNALMo significantly outperformed Graph Matching. The full statistical comparisons are provided in Appendix~\ref{app:cbbce-permutation-histograms}.

\subsection{SDSM Model Results}
\begin{table}[H]
\centering
\caption{Mean F1 score by RNA family for each structure extraction method using the SDSM model.}
\label{tab:f1-by-family-sdsm}
\begin{tabular}{lcccc}
\toprule
Family & RiNALMo F1 & Graph Matching F1 & Nussinov F1 & SPOT-RNA F1 \\
\midrule
5S rRNA  & 0.7631 & 0.9501 & 0.8899 & 0.9577 \\
SRP RNA  & 0.6151 & 0.7383 & 0.7467 & 0.7489 \\
tRNA     & 0.7111 & 0.9215 & 0.8595 & 0.9219 \\
\midrule
\textbf{Mean} & \textbf{0.7171} & \textbf{0.8935} & \textbf{0.8494} & \textbf{0.9004} \\
\bottomrule
\end{tabular}
\end{table}

The SDSM model achieved its highest mean F1 scores with SPOT-RNA (0.9004) and Graph Matching extraction (0.8935), followed by Nussinov extraction (0.8494), while RiNALMo extraction produced the lowest score (0.7171), as shown in \Cref{tab:f1-by-family-sdsm}. The bootstrapping permutation tests showed that SPOT-RNA significantly outperformed all other extraction methods, Graph Matching significantly outperformed both Nussinov and RiNALMo extraction, and Nussinov extraction significantly outperformed RiNALMo extraction. Full pairwise statistical comparisons are reported in Appendix~\ref{app:sdsm-permutation-histograms}.

\section{Discussion}

\begin{table}[H]
\centering
\caption{Overall mean F1 score for each extraction method, and mean squared error between ground-truth target structure and direct model outputs (prior to structure extraction), across the three models trained in this work.}
\label{tab:mean-f1-by-model-and-extractor}
\begin{tabular}{lccccc}
\toprule
Model & RiNALMo F1 & Graph Matching F1 & Nussinov F1 & SPOT-RNA F1 & MSE \\
\midrule
Nussinov-like & 0.6524 & \textbf{0.9163} & 0.8451 & 0.6774 & 0.1459 \\
BCE (baseline) & 0.7147 & 0.6008 & 0.8478 & \textbf{0.8707} & 0.00196 \\
SDSM & 0.7171 & 0.8935 & 0.8494 & \textbf{0.9004} & 0.000496 \\
\bottomrule
\end{tabular}
\end{table}

Our results show that training-extraction congruence plays an important role in RNA secondary structure prediction. The pretrained RiNALMo model provides a fixed-model reference: the model and its output matrices were held constant while only the extraction method was varied. All four extractors achieved similar performance, indicating that these outputs were relatively extractor-agnostic and that changing the extraction method alone was unlikely to yield substantial performance gains in this case. This experiment is separate from the proof-of-concept models trained in this paper and the comparisons that follow. Thus, the absolute scores of the varying extractors on predictions from the pre-trained RiNALMo model are not directly comparable to those of the models trained in this paper as RiNALMo differs substantially in architecture, training data, and optimization.

In contrast to the relatively extractor-agnostic RiNALMo outputs, the Nussinov-like model showed a clear dependence on extraction method. The two greedy extractors performed worst, with SPOT-RNA at 0.6774 and RiNALMo at 0.6524, likely due to the fact that threshold-based greedy extraction is incongruent to the pseudo-free energy weights this model produces, even after those weights are transformed into the expected ranges. The congruent Nussinov extractor performed well (0.8451), supporting the central claim of this work that aligning the training objective with the extractor improves performance. However, Graph Matching performed best overall (0.9163), which is also the single highest result across every model and extractor in this work. While Graph Matching is not necessarily more congruent with the Nussinov-like model training objective than the Nussinov-like extractor is, it is generally a stronger extractor method in this setting as the Nussinov-like model produces additive pseudo-free energies, which are exactly the linearly comparable weights that Graph Matching assumes. So by guaranteeing to select the maximum-weight matching it will likely select the strongest structure. Also, unlike the Nussinov extractor, Graph Matching is not restricted to nested structures, so when the learned matrix contains pseudoknots, it is likely to extract a better discrete structure.

The binary cross-entropy (BCE) baseline model produced a very different ranking. SPOT-RNA performed best (0.8707), the Nussinov extractor and RiNALMo extractor occupied an intermediate range (0.8478 and 0.7147), and Graph Matching performed worst by a wide margin (0.6008). The collapse of Graph Matching is the most informative result here, because it exposes a mismatch between this model's outputs and what Graph Matching assumes. Graph Matching treats the input scores as linearly comparable additive weights. The BCE model, however, produces bounded sigmoid confidences that are computed independently for each entry and have no such additive interpretation. A further problem comes from the treatment of unpaired states. Leaving two bases unpaired contributes both diagonal terms, $a_{ii}$ and $a_{jj}$, whereas pairing them contributes only the single off-diagonal term $a_{ij}$, so a pair is selected only when $a_{ij} > a_{ii} + a_{jj}$. Because the sigmoid entries are not normalized, the single pairing score and the two unpaired scores are not on a comparable scale. Together, these mismatches explain why Graph Matching performs so poorly on a model whose outputs it cannot interpret as additive weights, despite performing best on the Nussinov model.

The SDSM model resolves this training-extraction congruence problem directly. It shares the architecture and loss of the baseline BCE model and differs only in the SDSM normalization applied to its output, yet it outperformed the baseline on every extraction method, with all four improvements statistically significant. The largest gain was for Graph Matching, which rose from 0.6008 to 0.8935. This recovery follows directly from the analysis above. SDSM normalization forces each base's pairing weights and its unpaired weight to sum to one, so the diagonal and off-diagonal entries become parts of a single probability distribution and are directly comparable. SPOT-RNA achieved the highest score on the SDSM model (0.9004), with Graph Matching close behind (0.8935). RiNALMo was among the weaker extractors across the trained models, and remained the lowest-performing extractor for the SDSM model (0.7171). This is likely because its default 0.5 confidence threshold discards pairs whose probability mass is spread below the cutoff once the matrix is normalized, where SPOT-RNA's lower threshold retains more of them.

\begin{figure}[H]
\centering
\includegraphics[width=\textwidth]{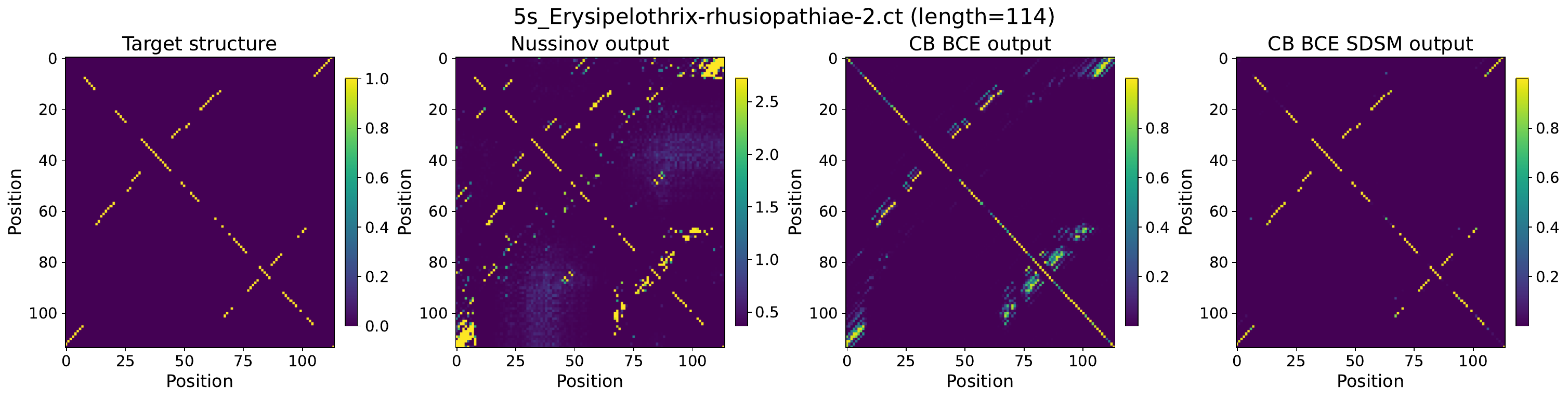}
\caption{Predicted matrices for a representative sequence (5S\_Erysipelothrix-rhusiopathiae-2, length 114): target structure, Nussinov-like output, CB BCE output, and CB BCE SDSM output. The SDSM output concentrates probability mass onto the true base pairs with little off-target signal.}
\label{fig:model-output-comparison}
\end{figure}

Two further properties of the SDSM model are worth noting. First, it produced the most consistent performance across extractors. Every extraction method scored above 0.71, giving a spread of 0.18 compared with 0.26 for the Nussinov model and 0.27 for the baseline. Because the SDSM output is already an interpretable structure representation, the choice of extractor matters far less, which is the practical payoff of producing a probability matrix directly rather than relying on a downstream extractor. Second, it achieved by far the lowest mean squared error of the three models (0.000496, against 0.00196 for the baseline and 0.1459 for the Nussinov model), indicating that its output is closest to the ground truth before any extraction is applied. This is visible in \Cref{fig:model-output-comparison}, where the SDSM output concentrates its probability mass tightly onto the true base pairs and shows little off-target signal, while the Nussinov and baseline outputs are more diffused.

\section{Conclusion}

This work demonstrates the importance of training-extraction congruence in RNA secondary structure prediction. When congruence is low, a model can minimize its training loss while producing base-pairing weight matrices that are poorly suited to extraction into interpretable representations. These interpretable representations are often required for downstream tasks and come in the form of a single discrete structure or a probability matrix.

In looking at the results for the accuracy of the structures predicted directly by each model and the accuracy post extraction, our results show that model training affects both the quality of the output representation and how effectively it can be extracted. The substantial reshuffling of extractor rankings between the Nussinov-like and binary cross-entropy (BCE) models demonstrates that no extractor is universally optimal. Instead, its effectiveness depends on how closely its procedure and assumptions align with those used during model training.

The symmetric doubly stochastic matrix (SDSM) model addresses this training-extraction congruence problem directly. The differentiable SDSM normalization used in this model allows it to produce a non-negative, symmetric, doubly stochastic base-pairing probability matrix directly, removing the need for the additional structure extraction for almost all downstream tasks. The SDSM model achieved the lowest pre-extraction error and outperformed the BCE baseline under every evaluated extraction method.

Taken together, these results suggest two approaches to improving RNA secondary structure prediction: align the training objective with the intended extraction method, or train the model to produce the required representation directly. SDSM normalization demonstrates the potential of the second approach by producing an interpretable probability matrix while retaining the option of discrete extraction when required. Although the models trained here were intentionally small proof-of-concept models, applying SDSM normalization to larger architectures and evaluating it under inter-family generalization settings are promising directions for future work.

\section{References}
\label{sec:ref}
\printbibliography[heading=none]

\appendix

\section*{Appendix}

\section{Model Architectures}
\label{app:model-architectures}

\begin{table}[H]
\centering
\caption{Architecture for the fully connected MLP models used in this work. All models share the same base structure, consisting of three ReLU-activated hidden layers followed by a dense output layer of size $200 \times 200$.}
\label{tab:all-model-architectures}
\begin{tabular}{@{}lccc@{}}
\toprule
\textbf{Layer} & \textbf{Type} & \textbf{Neurons} & \textbf{Activation} \\
\midrule
1 & Input  & 800 ($200 \times 4$)        & -- \\
2 & Dense  & 1024                         & ReLU \\
3 & Dense  & 512                          & ReLU \\
4 & Dense  & 256                          & ReLU \\
5 & Dense  & 40{,}000 ($200 \times 200$)  & $^{\ast}$ \\
\bottomrule
\end{tabular}

\vspace{0.5em}
\footnotesize{$^{\ast}$Final layer activation differs by model: \textit{tanh} for the Nussinov-like model; \textit{sigmoid} for the CB BCE baseline; raw scores followed by softplus and SDSM normalization for the SDSM model.}
\end{table}

\section{Statistical Significance Testing}

\subsection{Pre-trained RiNALMo Model: Pairwise Extraction Method Comparisons}
\label{app:rinalmo-permutation-histograms}

\begin{table}[H]
\centering
\caption{Pairwise permutation test $p$-values for the pre-trained RiNALMo model outputs on the full ArchiveII dataset ($n=3{,}975$; 10{,}000 permutations). Each cell gives the $p$-value for the hypothesis that the \textbf{row} method outperforms the \textbf{column} method. Significant results ($p < 0.05$) are shown in bold.}
\label{tab:rinalmo-pairwise-pvals}
\begin{tabular}{lcccc}
\toprule
\textbf{Row $>$ Column} & \textbf{Graph Matching} & \textbf{Nussinov} & \textbf{RiNALMo} & \textbf{SPOT-RNA} \\
\midrule
Graph Matching  & --                  & $p=1.000$          & $p=0.999$          & $p=1.000$ \\
Nussinov        & \textbf{$<0.001$}   & --                 & $p=0.127$          & \textbf{$<0.001$} \\
RiNALMo         & \textbf{$<0.001$}   & $p=0.873$          & --                 & \textbf{$<0.001$} \\
SPOT-RNA        & \textbf{$<0.001$}   & $p=0.999$          & $p=0.999$          & -- \\
\bottomrule
\end{tabular}
\end{table}

\subsection{Nussinov-like Model: Pairwise Extraction Method Comparisons}
\label{app:nussinov-permutation-histograms}

\begin{table}[H]
\centering
\caption{Pairwise permutation test $p$-values for the Nussinov-like model (10{,}000 permutations). Each cell gives the $p$-value for the hypothesis that the \textbf{row} method outperforms the \textbf{column} method. Significant results ($p < 0.05$) are shown in bold.}
\label{tab:nussinov-pairwise-pvals}
\begin{tabular}{lcccc}
\toprule
\textbf{Row $>$ Column} & \textbf{Graph Matching} & \textbf{Nussinov} & \textbf{RiNALMo} & \textbf{SPOT-RNA} \\
\midrule
Graph Matching  & --                  & \textbf{$<0.001$}  & \textbf{$<0.001$}  & \textbf{$<0.001$} \\
Nussinov        & $p=1.000$           & --                 & \textbf{$<0.001$}  & \textbf{$<0.001$} \\
RiNALMo         & $p=1.000$           & $p=1.000$          & --                 & $p=0.999$ \\
SPOT-RNA        & $p=1.000$           & $p=1.000$          & \textbf{$p=0.001$} & -- \\
\bottomrule
\end{tabular}
\end{table}

\subsection{CB BCE Baseline Model: Pairwise Extraction Method Comparisons}
\label{app:cbbce-permutation-histograms}

\begin{table}[H]
\centering
\caption{Pairwise permutation test $p$-values for the CB BCE baseline model (10{,}000 permutations). Each cell gives the $p$-value for the hypothesis that the \textbf{row} method outperforms the \textbf{column} method. Significant results ($p < 0.05$) are shown in bold.}
\label{tab:cbbce-pairwise-pvals}
\begin{tabular}{lcccc}
\toprule
\textbf{Row $>$ Column} & \textbf{Graph Matching} & \textbf{Nussinov} & \textbf{RiNALMo} & \textbf{SPOT-RNA} \\
\midrule
Graph Matching  & --                   & $p=1.000$          & $p=0.999$          & $p=1.000$ \\
Nussinov        & \textbf{$<0.001$}    & --                 & \textbf{$<0.001$}  & $p=1.000$ \\
RiNALMo         & \textbf{$p=0.001$}   & $p=1.000$          & --                 & $p=1.000$ \\
SPOT-RNA        & \textbf{$<0.001$}    & \textbf{$<0.001$}  & \textbf{$<0.001$}  & -- \\
\bottomrule
\end{tabular}
\end{table}

\subsection{SDSM Model: Pairwise Extraction Method Comparisons}
\label{app:sdsm-permutation-histograms}

\begin{table}[H]
\centering
\caption{Pairwise permutation test $p$-values for the SDSM model (10{,}000 permutations). Each cell gives the $p$-value for the hypothesis that the \textbf{row} method outperforms the \textbf{column} method. Significant results ($p < 0.05$) are shown in bold.}
\label{tab:sdsm-pairwise-pvals}
\begin{tabular}{lcccc}
\toprule
\textbf{Row $>$ Column} & \textbf{Graph Matching} & \textbf{Nussinov} & \textbf{RiNALMo} & \textbf{SPOT-RNA} \\
\midrule
Graph Matching  & --                    & \textbf{$<0.001$}  & \textbf{$<0.001$}  & $p=1.000$ \\
Nussinov        & $p=1.000$             & --                 & \textbf{$<0.001$}  & $p=1.000$ \\
RiNALMo         & $p=1.000$             & $p=1.000$          & --                 & $p=1.000$ \\
SPOT-RNA        & \textbf{$p=0.0002$}   & \textbf{$<0.001$}  & \textbf{$<0.001$}  & -- \\
\bottomrule
\end{tabular}
\end{table}

\subsection{Nussinov-like Model vs CB BCE Baseline}
\label{app:model-vs-model-nussinov}

\begin{table}[H]
\centering
\caption{Model-vs-model permutation test $p$-values comparing the Nussinov-like model against the CB BCE baseline, per extraction method. Significant results ($p < 0.05$) are shown in bold.}
\label{tab:model-vs-model-nussinov-cbbce}
\begin{tabular}{lcccc}
\toprule
\textbf{Extraction Method} & \textbf{Mean Nussinov F1} & \textbf{Mean CB BCE F1} & \textbf{Nussinov $>$ CB BCE} & \textbf{CB BCE $>$ Nussinov} \\
\midrule
Graph Matching  & 0.9163 & 0.6008 & \textbf{$<0.001$} & $p=1.000$ \\
Nussinov        & 0.8451 & 0.8478 & $p=1.000$         & \textbf{$<0.001$} \\
RiNALMo         & 0.6524 & 0.7147 & $p=1.000$         & \textbf{$<0.001$} \\
SPOT-RNA        & 0.6774 & 0.8707 & $p=1.000$         & \textbf{$<0.001$} \\
\bottomrule
\end{tabular}
\end{table}

\subsection{SDSM Model vs CB BCE Baseline}
\label{app:model-vs-model-sdsm}

\begin{table}[H]
\centering
\caption{Model-vs-model permutation test $p$-values comparing the SDSM model against the CB BCE baseline, per extraction method. Significant results ($p < 0.05$) are shown in bold.}
\label{tab:model-vs-model-sdsm-cbbce}
\begin{tabular}{lcccc}
\toprule
\textbf{Extraction Method} & \textbf{Mean SDSM F1} & \textbf{Mean CB BCE F1} & \textbf{SDSM $>$ CB BCE} & \textbf{CB BCE $>$ SDSM} \\
\midrule
Graph Matching  & 0.8935 & 0.6008 & \textbf{$<0.001$}  & $p=1.000$ \\
Nussinov        & 0.8494 & 0.8478 & \textbf{$<0.001$}  & $p=1.000$ \\
RiNALMo         & 0.7171 & 0.7147 & \textbf{$p=0.001$} & $p=1.000$ \\
SPOT-RNA        & 0.9004 & 0.8707 & \textbf{$<0.001$}  & $p=1.000$ \\
\bottomrule
\end{tabular}
\end{table}

\section{SDSM Normalization Implementation}
\label{app:sdsm-code}
\begin{algorithm}[H]
\caption{Modified Sinkhorn--Knopp normalization}
\label{alg:sdsm-normalization}

\Fn{\textsc{NormalizeSDSM}%
    $(A \in \mathbb{R}_{>0}^{L \times L},
      T=100,\tau=10^{-8},\varepsilon=10^{-8})$}{

    $X \leftarrow A$\;
    $\mathrm{converged} \leftarrow \mathrm{False}$\;

    \Comment{Iteratively normalize while preserving symmetry}
    \For{$t \leftarrow 1$ \KwTo $T$}{

        $\widetilde{X}_{ij}
        \leftarrow
        \dfrac{X_{ij}}
        {\sum_k X_{kj}+\varepsilon}$
        \Comment*[r]{normalize columns}

        $\widetilde{X}_{ij}
        \leftarrow
        \dfrac{\widetilde{X}_{ij}}
        {\sum_k \widetilde{X}_{ik}+\varepsilon}$
        \Comment*[r]{normalize rows}

        $\widetilde{X}
        \leftarrow
        \dfrac{1}{2}
        \left(\widetilde{X}+\widetilde{X}^{\top}\right)$
        \Comment*[r]{restore symmetry}

        $e_{\mathrm{row}}
        \leftarrow
        \max_i\left|
        \sum_j\widetilde{X}_{ij}-1
        \right|$\;

        $e_{\mathrm{col}}
        \leftarrow
        \max_j\left|
        \sum_i\widetilde{X}_{ij}-1
        \right|$\;

        \If{$\neg\,\mathrm{converged}$}{
            $X \leftarrow \widetilde{X}$\;
            $\mathrm{converged}
            \leftarrow
            (e_{\mathrm{row}}<\tau)
            \land
            (e_{\mathrm{col}}<\tau)$\;
        }
    }

    \KwRet $X$\;
}
\end{algorithm}

\end{document}